\documentclass{article}
\usepackage{ijcai19}

\usepackage{times}
\usepackage{soul}
\usepackage{url}
\usepackage[utf8]{inputenc}
\usepackage[small]{caption}
\usepackage{graphicx}
\usepackage{amsmath}
\usepackage{booktabs}
\usepackage{algorithm}
\usepackage{algorithmic}

\usepackage{amsmath}

\usepackage{xcolor}
\usepackage{multirow}
\usepackage{booktabs}
\usepackage{multirow}
\usepackage{placeins}
\usepackage{tabularx}
\usepackage{threeparttable}

\usepackage{amsfonts}
\usepackage{microtype}
\usepackage{graphicx}
\usepackage{booktabs} 
\usepackage{adjustbox}
\usepackage{subcaption}

\def\Softmax{\textsf{Softmax}}

\title{Bayesian Uncertainty Matching for Unsupervised Domain Adaptation}

\author{
Jun Wen $^{1,2}$
\and
Nenggan Zheng$^{1,2}$\thanks{Corresponding author.}\and
Junsong Yuan$^3$\and
Zhefeng Gong$^{4}$\And
Changyou Chen$^3$
\affiliations
$^1$Qiushi Academy for Advanced Studies, Zhejiang University, Hangzhou, China\\
$^2$College of Computer Science and Technology, Zhejiang University, Hangzhou, China\\
$^3$Computer Science and Engineering Department, State University of New York at Buffalo \\
$^4$Department of Neurobiology, Zhejiang University School of Medicine, Hangzhou, China
\emails
\{junwen,zfgong\}@zju.edu.cn,
zng@cs.zju.edu.cn,
\{jsyuan,changyou\}@buffalo.edu
}

\begin{document}

\maketitle

\begin{abstract}
Domain adaptation is an important technique to alleviate performance degradation caused by domain shift, e.g., when training and test data come from different domains. Most existing deep adaptation methods focus on reducing domain shift by matching marginal feature distributions through deep transformations on the input features, due to the unavailability of target domain labels. We show that domain shift may still exist via label distribution shift at the classifier, thus deteriorating model performances. To alleviate this issue, we propose an approximate joint distribution matching scheme by exploiting prediction uncertainty. Specifically, we use a Bayesian neural network to quantify prediction uncertainty of a classifier. By imposing distribution matching on both features and labels (via uncertainty), label distribution mismatching in source and target data is effectively alleviated, encouraging the classifier to produce consistent predictions across domains. We also propose a few techniques to improve our method by adaptively reweighting domain adaptation loss to achieve nontrivial distribution matching and stable training. Comparisons with state of the art unsupervised domain adaptation methods on three popular benchmark datasets demonstrate the superiority of our approach, especially on the effectiveness of alleviating negative transfer.
\end{abstract}

\section{Introduction}
Many machine-learning algorithms assume that training and test data, typically in terms of feature-label pairs, denoted as $\{x_i, y_i\}_i$, are drawn from the same feature-label space with the same distribution, where $x_i$ is the feature while $y_i$ is the label of $x_i$. However, this assumption rarely holds in practice as the data distribution is likely to change over time and space. Though state-of-the-art deep convolutional features have shown invariant to low-level variations to some degree, they are still susceptible to domain-shift, as it is expensive to manually label sufficient training data that cover diverse application domains. A typical solution is to further finetune a learned deep model on task-specific datasets. However, it is still prohibitively difficult and expensive to obtain enough labeled data for finetuning on a big deep network. Instead of re-collecting labeled data for every possible new task, unsupervised domain-adaptation methods are adopted to alleviate performance degradations by transferring knowledge from related labeled source domains to an unlabeled target domain \cite{ganin2016domain,li2017end,JMLR:v20:13-580}.

\begin{figure}[t!]
\begin{center}
\centerline{\includegraphics[width=0.7\columnwidth]{{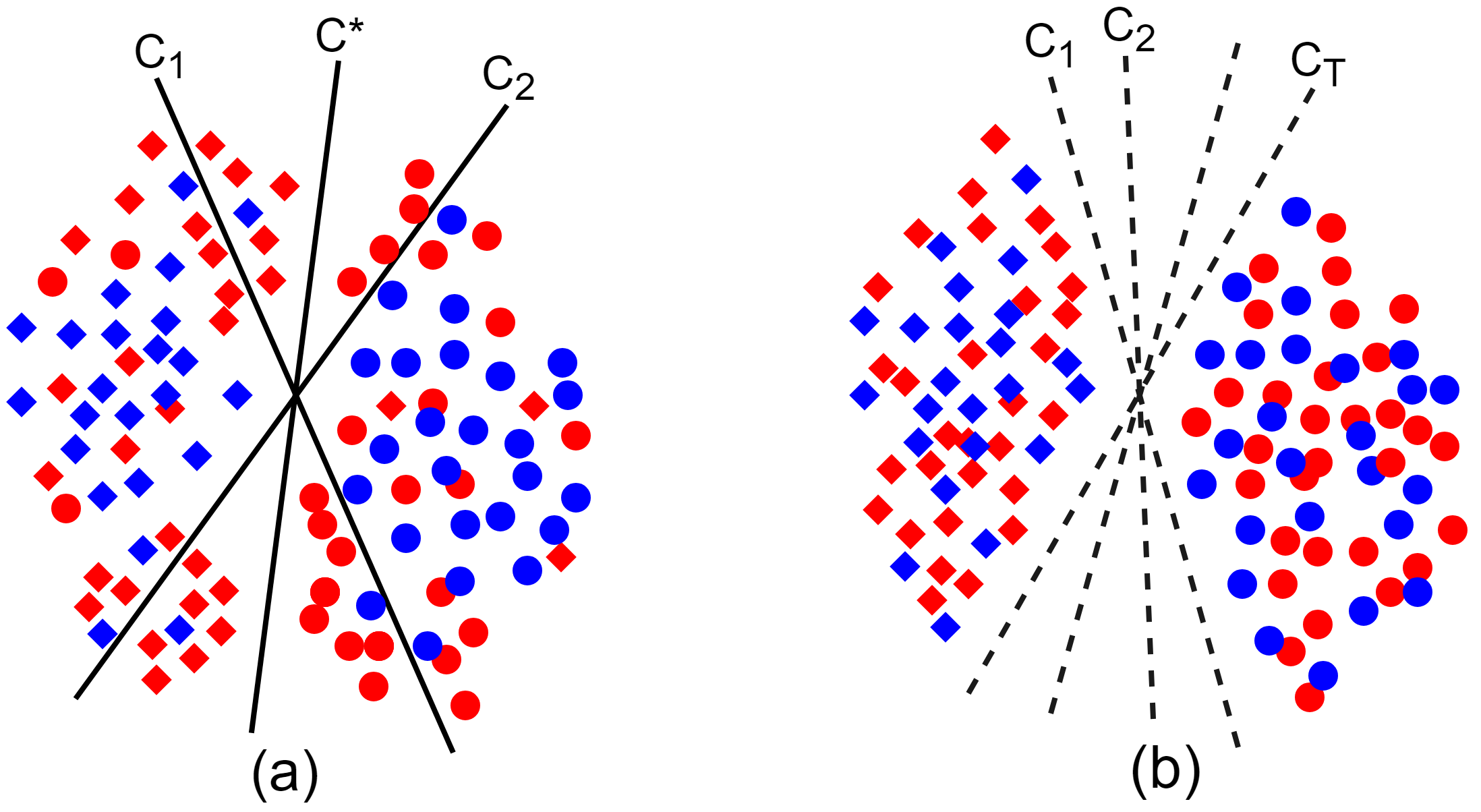}}}
\caption{Comparisons between conventional and the proposed domain-adaptation methods (blue: source domain and red: target domain; diamonds and circles are samples from two different categories). Standard methods reduce domain-shift through {\em marginal feature-distribution} matching, which could learn a source-biased classifier $C_1$ or $C_2$ when the label-distributions do not match ($C^*$ denotes a domain-invariant classifier). b) Our method jointly matches feature-distributions and label-distributions by learning domain-consistent probabilistic classifiers (sampled as $C_1,C_2,...,C_T$) with uncertainty matching.}\label{fig:compare}
\end{center}
\vspace{-0.8cm}
\end{figure}


When adopting domain adaptation, certain assumptions must be imposed on how distributions change across domains. For instance, most existing domain adaptation methods consider a covariate shift situation where the distributions on source and target domains only differ in the marginal feature-distribution $P(X)$, with an identical conditional distribution $P(Y |X)$ assumption. Here we use $X$ and $Y$ to denote random variables whose realizations are features $x_i$ and labels $y_i$, either from the source data $(X_s, Y_s)$ or target data $(X_t, Y_t)$. In this setting, an early attempt is to match the feature distribution $P(X)$ on source and target domains by importance reweighting \cite{huang2007correcting}. State-of-the-art approaches reduce domain-shift by learning domain-invariant representations through deep neural transformations $\textbf{G}_{\phi}(X)$, parameterized by $\phi$, such that $P(\textbf{G}_{\phi}(X_s)) \approx P(\textbf{G}_{\phi}(X_t))$. This is often achieved by optimizing a deep network to minimize some distribution-discrepancy measures \cite{sun2016deep,tzeng2017adversarial}. Because there is no target-domain label in the unsupervised domain adaptation scenario, most existing methods simply assume $P(Y_t| (\textbf{G}_{\phi}(x_t))) \approx P (Y_s| (\textbf{G}_{\phi}(x_s)))$ by sharing a classifier learned with source labeled data only. However, this is typically not true in practice as the source-learned classifier tends to be biased toward the source. As shown in Figure~\ref{fig:compare} (a), though the feature-distributions are well matched, the classifiers may still perform poorly in the target domain due to label-distribution mismatch.

In this paper, we alleviate the above problem by proposing an approximate joint-distribution matching scheme. Specifically, due to the lack of label information in a target domain, we propose to match the model prediction uncertainty, a second-order statistic equivalent, induced by the conditional distribution $P(Y|\textbf{G}_{\phi}(X))$. We obtain the prediction uncertainty by imposing a Bayesian neural network (BNN) which induces posterior distributions over weights of a neural network. Without uncertainty matching, the BNN classifier is expected to produce high uncertainty for the unseen target-domain data and low uncertainty for the source-domain data, due to the bias induced by training on the source domain. By contrast, with prediction uncertainty matching, one is able to achieve an approximate joint-distribution matching, alleviating domain-shift on the classifier. The contributions of our work are summarized as follows:

\begin{itemize}

\item Different from most existing domain-adaptation methods, which only focus on reducing marginal feature-distribution discrepancy, we propose to match joint {\em feature-label distributions} by exploiting model prediction uncertainty, effectively alleviating conditional-distribution shift imposed by the classifier.

\item We employ BNNs to quantify prediction uncertainty. Through additional source and target uncertainty discrepancy minimization, both fine-grained marginal feature-distribution and conditional label-distribution matching are achieved.

\item  Extensive experimental results on standard domain-adaptation benchmarks demonstrate the effectiveness of the proposed method, outperforming current state-of-the-art approaches.

\end{itemize}

\section{Related Works}


\subsection{Domain Adaptation}
Domain adaptation methods seek to learn discriminative features from neighbouring source domains to target domains. This is usually achieved by learning domain-invariant features \cite{ben2010theory}. Previous methods usually seek to align source and target feature through subspace learning \cite{gong2012geodesic}. Recently, deep adversarial-domain-adaptation approaches have taken over and achieved state-of-the-art performances \cite{pmlr-v80-hoffman18a,wen2019exploiting}. These methods attempt to reduce domain discrepancy by optimizing deep networks with an adversarial objective produced by a discriminator network that is trained to distinguish features of target from source domains. Though significant marginal distribution-shift can be reduced, these methods fail to fully address the conditional label-distribution shift problem. There are some recent models trying to address this issue by utilizing pseudo-labels \cite{long2018conditional,chen2018re}. However, most of them are deterministic models, which can not essentially reduce the conditional domain-shift, due to the unavailability of target-domain labels.

\subsection{Bayesian Uncertainty}
Uncertainty can be achieved by adopting Bayesian neural networks. A typical BNN assigns a prior distribution, {\it e.g.}, a Gaussian prior distribution, over the weights, instead of deterministic weights as in standard neural networks. Given observed data, approximate inference is performed to calculate posterior distribution of the weights, such as the methods in \cite{graves2011practical,blundell2015weight}. A more effective way to calculate Bayesian uncertainty is to employ the dropout variational inference \cite{gal2016dropout}, which is adopted in this paper.

\section{The Proposed Method}

\subsection{The Overall Idea}

Given a labeled source-domain dataset $D_s=(X_s,Y_s)$ and an unlabeled target-domain dataset $D_t=(X_t$), the goal of unsupervised domain-adaptation is to learn an adapted model from the labeled source-domain data to the unlabeled target-domain data. The source and target domains are assumed to be sampled from two joint distributions $P_s(X_s,Y_s)$ and $P_t(X_t,Y_t)$, respectively, with $P_s \ne P_t$. The joint distribution of feature-label pairs can be decomposed as:
\begin{equation}
\small
P(X,Y)=P(Y|X)P(X).
\end{equation}

\paragraph{Limitations of Traditional Methods.}
Most existing domain-adaptation methods reduce domain-shift by learning a deep feature-transformation $\textbf{G}_{\phi}$ such that $P (\textbf{G}_{\phi}(X_s)) \approx P(\textbf{G}_{\phi}(X_t))$, and a shared classifier network $P_{\theta}(Y_s|\textbf{G}_{\phi}(X_s))$, parameterized by $\theta$, using labeled source data $D_s$. To adapt to a target domain, the learned $P_{\theta}(Y_s|\textbf{G}_{\phi}(X_s))$ is adopted to form the target-domain joint distribution $P(\textbf{G}_{\phi}(X_t))P_{\theta}(Y_t|\textbf{G}_{\phi}(X_t))$. It is easy to see that directly adopting $P_{\theta}(Y_s|\textbf{G}_{\phi}(X_s))$ in the target-domain is unable to match the true joint distributions $P_s(X_s,Y_s)$ and $P_t(X_t,Y_t)$, as $P_{\theta}(Y_s|\textbf{G}_{\phi}(X_s))$ only reflects feature-label information in the source domain.

\subsubsection{Our Method}
In this paper, we propose to jointly reduce the marginal-distribution shift ($P(X_s) \ne P(X_t)$) and conditional-distribution shift ($P(Y_s|X_s) \ne P(Y_t|X_t)$) by exploiting prediction uncertainty. Specifically, our model consists of a probabilistic BNN feature extractor $\textbf{G}_{\phi}$ with inputs $X_s$ or $X_t$, and a BNN classifier $\textbf{C}_{\theta}$ with inputs $\textbf{G}_{\phi}(X_s)$ or $\textbf{G}_{\phi}(X_t)$. The classifier $\textbf{C}_{\theta}$, which corresponds to the conditional distribution $P(Y|\textbf{G}_{\phi}(X))$ and is parameterized by $\theta$, learns to classify samples from both domains.

\begin{figure}[ht]
\begin{center}
\centerline{\includegraphics[width=0.8\columnwidth]{{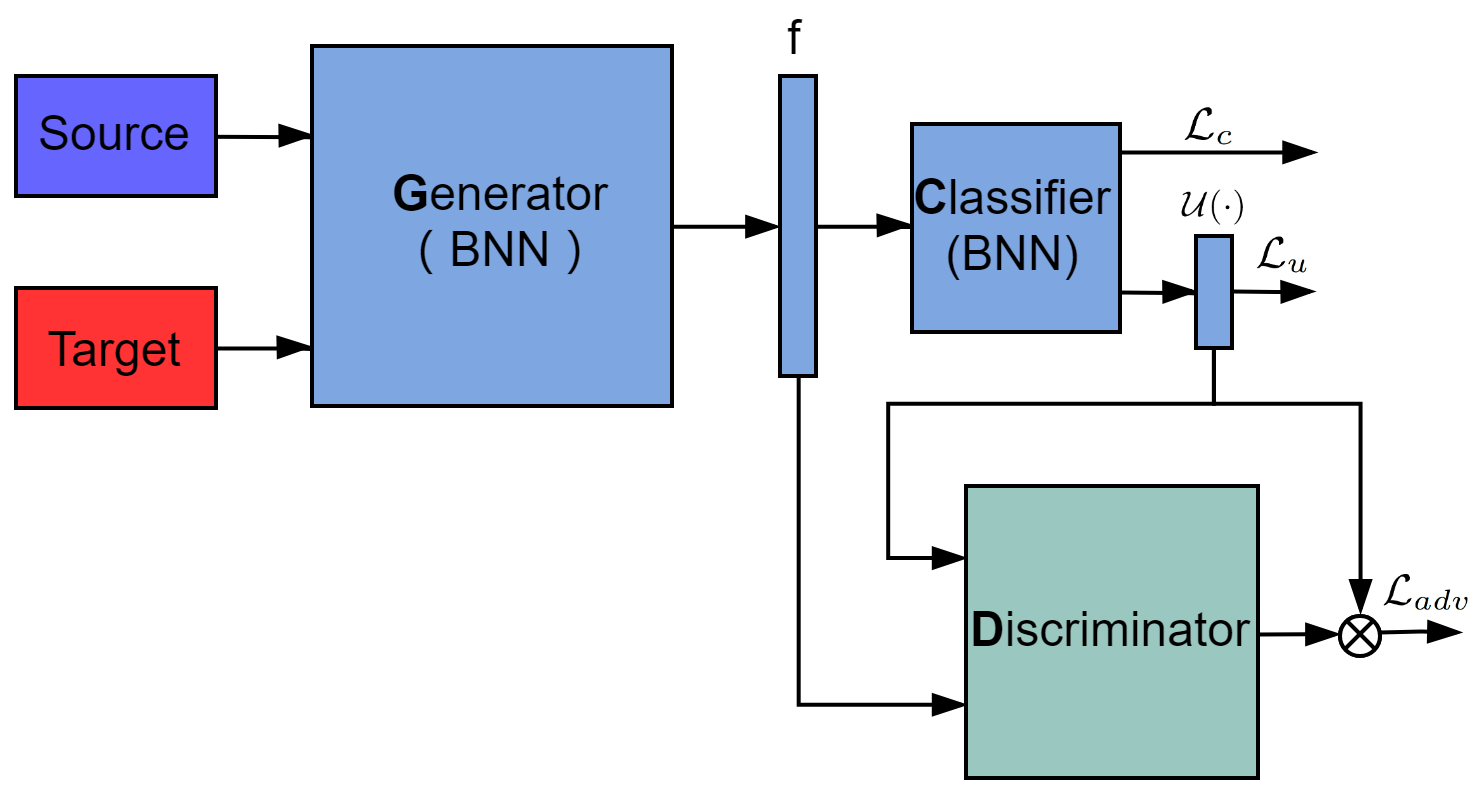}}}
\caption{Pipeline of the proposed method. We adaptively match the joint distribution of the learned feature and prediction uncertainty.}
\label{fig:pipeline}
\end{center}
\vspace{-0.8cm}
\end{figure}

As discussed in the Introduction, directly learning to match $P(Y_s|\textbf{G}_{\phi}(X_s))$ and $P(Y_t|\textbf{G}_{\phi}(X_t))$ is unfeasible due to the unavailability of target labels. To overcome the difficulty, we instead learn to match the prediction uncertainty, a second-order statistics equivalent. The intuition is that if the second-order statistics of two distributions are matched, the two distributions will be brought closer. Another intuition is that, if target samples are not well matched with source samples in the feature space, these outliers are likely to be predicted with high uncertainty by a source-trained classifier. If one can quantify the uncertainty and minimize the cross-domain uncertainty discrepancy (source uncertainty is supposed to be low), the generator $\textbf{G}_{\phi}$ will be encouraged to produce target features that best match the source both in the feature space and classifier prediction. In the following, we first introduce an effective way to obtain Bayesian uncertainty by adopting the dropout technique, and then describe the proposed framework of joint-distribution matching.

\subsection{Bayesian Uncertainty}
We employ Bayesian neural network (BNN) to quantify model prediction uncertainty. BNN is a variant of standard neural networks by treating the weights as distributions, instead of using deterministic weights. However, it is often computationally inhibited to perform inference on the weight distributions in a large-scale deep BNN. In this paper, we employ the practical dropout variational inference for approximate inference \cite{gal2016dropout} and efficient uncertainty approximation. In the proposed method, inference is done by training the model with dropout \cite{srivastava2014dropout}. In testing, dropout is also performed to generate approximate samples from the posterior distribution. This approach is equivalent to using a Bernoulli variational distribution $q_\vartheta(\textbf{W})$ \cite{gal2016dropout}, parameterized by $\vartheta$, to approximate the true model weights ($\textbf{W}$) posterior. As proven in \cite{gal2016dropout}, the dropout inference essentially minimizes the KL divergence between the approximate distribution and the posterior of a deep Gaussian process. For classification, the objective can be formulated as:
\begin{equation}
\small
\mathcal{L}_{(\theta,p)} = -\frac{1}{N}\sum_{i=1}^{N} \log p(y_i|f^{\hat{\textbf{W}}_i}(x_i))+\frac{1-p}{2N}||\vartheta||^2,
\end{equation}
where $N$ is the number of training samples, $p$ denotes the dropout probability, $\hat{\textbf{W}}_i $ is sampled according to the dropout variational distribution $q^*_\vartheta(\textbf{W})$ \cite{gal2016dropout}, and $\vartheta$ is the set of the variational distribution's parameters.

The final prediction can be obtained by marginalizing over the approximate posterior distribution on weights, which is approximated using Monte Carlo integration as follows:
\begin{equation}
\small
p(y_i=c|x_i,X,Y)=\frac{1}{T}\sum_{t=1}^{T}\Softmax(f^{\hat{\textbf{W}}_t}(x_i)),
\end{equation}
\noindent with $T$ sampled masked weights, namely forwarding each sample $x_i$ through the feature extractor $\textbf{G}_{\phi}$ and classifier $\textbf{C}_{\theta}$ for $T$ times with weights sampled according to the dropout inference. The uncertainty of the prediction can be summarized using different metrics. In this paper, we use two metrics: 1) entropy of the averaged probabilistic prediction, and 2) variance of all prediction vectors. The entropy and variance based prediction uncertainty are denoted as $\mathcal{U}_{entro}$  and $\mathcal{U}_{var}$, respectively, formulated as:
\begin{equation}
\small
\mathcal{U}_{entro}(x_i) = H(\frac{1}{T}\sum_{t=1}^{T}\Softmax(\textbf{C}_{\theta}(\textbf{G}_{\phi}(x_i))/\tau)),
\end{equation}

\begin{equation}
\small
\begin{split}
\mathcal{U}_{var}(x_i) = \frac{1}{T} \sum_{t=1}^{T}(\textbf{C}_{\theta}(\textbf{G}_{\phi}(x_i)) - \frac{1}{T} \sum_{t=1}^{T} \textbf{C}_{\theta}(\textbf{G}_{\phi}(x_i)))^2,
\end{split}
\end{equation}

\noindent where $H(\cdot)$ denotes the information entropy function and $\tau$ the temperature of the $Softmax$, which controls the uncertainty level.

\subsection{Distribution Adaptation}
In this section, we describe how to simultaneously alleviate the marginal and conditional domain-shift by matching the approximate joint distributions of the source and target feature-label pairs.

\subsubsection{Joint-Distribution Adaptation}
We employ adversarial learning to match source and target statistics to reduce distribution discrepancy, as adversarial domain-adaptation methods have achieved state-of-the-art performances \cite{goodfellow2014generative,tzeng2017adversarial}. Basically, the procedure is described by a two-player game. The first player, a domain discriminator $\textbf{D}$, is trained to distinguish source from target data; while the second player, the feature extractor $\textbf{G}_{\phi}$, is trained to learn features that confuse the domain discriminator. By learning a best possible discriminator, the feature extractor is expected to learn features that are best domain-invariant. This learning procedure can be described by the following \emph{minimax} game:
\begin{equation}
\small
\begin{split}
\min \limits_{\textbf{G}_{\phi}} \max \limits_{\textbf{D}} \mathcal{L}_{adv} = -\frac{1}{n_s}\sum_{i=1}^{n_s}(\log (\textbf{D}(\textbf{G}_{\phi}(x_i^s)))) \\ -\frac{1}{n_t}\sum_{i=1}^{n_t}(\log (1-\textbf{D}(\textbf{G}_{\phi}(x_i^t)))),
\end{split}
\end{equation}
\noindent where $n_s$ and $n_t$ are the number of training samples from source and target, respectively.

However, this typical adversarial \emph{minimax} game for domain adaptation may be problematic in two aspects: 1) trivial feature alignment; and 2) unstable training. The domain discriminator fails to consider the relationship between learned features and the decision boundary of the classifier during feature alignment, which may lead to boundary target samples or trivial alignment with a huge-capacity $\textbf{G}_{\phi}$ \cite{shu2018dirt}. We aim to achieve nontrivial feature alignment by enforcing additional classifier prediction consistency during matching. Furthermore, noisy or hard-to-match samples may lead to unstable adversarial training. These confusing samples, which typically endow high prediction uncertainty, may produce unreliable gradients and deteriorate the training. They may also direct the $\textbf{G}_{\phi}$ to learn features that is non-discriminative for classifying target samples, especially with a huge-capacity $\textbf{G}_{\phi}$. Thus, we aim to attenuate the influence of noisy samples and reinforce the influence of easy-to-match target samples by adaptively re-weighting the adversarial loss. Specifically, we propose the following modified objective:

\begin{equation}
\small
\begin{split}
\min \limits_{\textbf{G}_{\phi}} \max \limits_{\textbf{D}} \mathcal{L}_{adv} = -\frac{1}{n_s}\sum_{i=1}^{n_s}(\alpha_{x_i^s}\log (\textbf{D}(\textbf{G}_{\phi}(x_i^s),\mathcal{U}(x_i^s)))) \\
-\frac{1}{n_t}\sum_{i=1}^{n_t}(\alpha_{x_i^t}\log (1-\textbf{D}(\textbf{G}_{\phi}(x_i^t),\mathcal{U}(x_i^t)))),
\end{split}
\end{equation}

\noindent where $\mathcal{U(\cdot)}$ is the prediction uncertainty formulated in Equation (4) or Equation (5). Both $\alpha_{x_i^s}$ and $\alpha_{x_i^t}$ are the adaptation loss weights, defined as:
\begin{equation}
 \alpha_{x_i}=
\begin{cases}
0 &  {\mathcal{U}(x_i) > t_u} \\
\frac{N \ast e^{-\mathcal{U}(x_i)}}{\sum_{i=1}^{N} e^{-\mathcal{U}(x_i)}}  & {\mathcal{U}(x_i)\leq t_u},
\end{cases}
\end{equation}

\noindent where $N$ is the number of training samples and $t_u$ denotes the uncertainty threshold constraining the influence of samples with uncertainty larger than $t_u$. For samples with uncertainty less than $t_u$, the weights are normalized within each training batch with more attention paid on the certain samples. It is worth noting that we found directly using the uncertainty without normalization for the re-weighting as done in \cite{kendall2017uncertainties,long2018conditional} tend to discourage a model from predicting low uncertainty for all samples. With such an adaptive joint-distribution adaptation objective, we aim to achieve non-trivial feature alignment and enable safer transfer.

\subsubsection{Conditional-Distribution Adaptation}
Note the joint-distribution-matching scheme described in the last section does not necessarily guarantee a good conditional-distribution adaptation.
In this section, we aim to reduce the conditional distribution shift and learn a domain-invariant classifier. Due to the infeasibility of directly minimizing the conditional distribution discrepancy $||P_\theta(Y|\textbf{G}_\phi(X_s))-P_\theta(Y|\textbf{G}_\phi(X_t))||_q$, we propose to approximate it by matching prediction uncertainty, a second-order statistic equivalent, with a BNN as the classifier. We exploit prediction uncertainty to detect and quantify domain-shift of a classifier. By minimizing the uncertainty discrepancy between source and target, we aim to approximately reduce the domain-shift of the classifier, and the objective $\mathcal{L}_u$ can be formulated as :
\begin{equation}
\small
\mathcal{L}_{u}= ||\mathcal{U}(X_s)-\mathcal{U}(X_t)||_q,
\end{equation}
\noindent where we set $q=2$ as we found it achieves better performances than $q=1$. The prediction uncertainty discrepancy is estimated within each batch during training.

To enable discriminative feature transferring, the feature extractor $\textbf{G}_{\phi}$ and classifier $\textbf{C}_{\theta}$ are also trained to minimize the source supervised loss $\mathcal{L}_c$ using source labels, defined as:
\begin{equation}
\small
\mathcal{L}_{c} = -\frac{1}{n_s}\sum_{i=1}^{n_s} y_{i}^s \cdot \log \Softmax (\textbf{C}_{\theta}(\textbf{G}_{\phi}(x_i^s))/\tau_c),
\end{equation}
\noindent where $ y_{i}^s$ is the true label of the source sample $x_i^s$ and $\tau_c$ is the $Softmax$ temperature for source classification.

Integrating all objectives together, the final learning procedure is formulated as:

\begin{equation}
\small
 \min \limits_{\textbf{G}_{\phi},\textbf{C}_{\theta}} \max \limits_{\textbf{D}} \mathcal{L}_{final}= \mathcal{L}_{c}+\lambda_{adv}\mathcal{L}_{adv}+\lambda_{u}\mathcal{L}_{u},
\end{equation}

\noindent where $\lambda_{adv}$ and  $\lambda_{u}$ are hyper-parameters that trade-off the objectives in the unified optimization problem.

According to the analysis of \cite{ben2010theory}, the expected target error is upper-bounded by the following three terms: 1) source error, 2) domain divergence, and 3) conditional-distribution discrepancy across domains. We aim to improve marginal distribution matching to reduce the second term by minimizing $\mathcal{L}_{adv}$ to achieve joint feature-uncertainty adaptation. While the third term is ignored by most of existing domain adaptation methods, we are able to reduce it via uncertainty matching and $\mathcal{L}_{u}$ minimization.

\section{Experiments}
We compare our method with state-of-the-art domain-adaptation approaches on several benchmark datasets: \emph{USPS-MNIST-SVHN} dataset \cite{pmlr-v80-hoffman18a}, \emph{Office-31} dataset \cite{saenko2010adapting}, and the recently introduced \emph{Office-home} dataset \cite{venkateswara2017deep}.

\paragraph{USPS-MNIST-SVHN.}
This dataset is used for digits recognition with 3 domains: MNIST, USPS, and SVNH. MNIST is composed of grey images of size $28\times28$; USPS contains $16\times16$ grey digits; and SVHN consists of $32\times32$ color digits images, which are more challenging and might contain more than one digit in each image. We evaluate our method using the three typical adaptation tasks: USPS$\leftrightarrow$MNIST (two tasks) and SVHN$\rightarrow$MNIST (one task). Following the same evaluation protocol of \cite{pmlr-v80-hoffman18a}, we use the standard training sets for domain-adaptation training and report adaptation results on the test sets.

\paragraph{Office-31.}
 This dataset is widely used for visual domain adaptation \cite{saenko2010adapting}. It consists of 4,652 images and 31 categories collected from three different domains: Amazon (A) from amazon.com, Webcam (W) and DSLR (D), taken by web camera and digital SLR camera in different environmental settings, respectively. We evaluate all methods on the following four challenging settings: A$\leftrightarrow$W and A$\leftrightarrow$D.

\paragraph{Office-home.}
 This is one of the most challenging visual domain adaptation datasets \cite{venkateswara2017deep}, which consists of 15,588 images with 65 categories of everyday objects in office and home settings. There are four significantly different domains: Art (Ar) consisting of 2427 painting, sketches or artistic depiction images, Clipart (Cl) containing 4365 images, Product (Pr) with 4439 images and Real-World (Rw) comprising of 4357 regularly captured images. We report performances of all the 12 adaptation tasks to enable thorough evaluations: Ar$\leftrightarrow$Cl, Ar$\leftrightarrow$Pr, Ar$\leftrightarrow$Rw, Cl$\leftrightarrow$Pr, Cl$\leftrightarrow$Rw, and Pr$\leftrightarrow$Rw.

\paragraph{Compared Methods.}
The state-of-the-art deep domain-adaptation methods we compared include: Domain Adversarial Neural Network (DANN) \cite{ganin2016domain}, Adversarial Discriminative Domain Adaptation (ADDA) \cite{tzeng2017adversarial}, Joint Adaptation Networks(JAN) \cite{long2017deep}, Conditional Domain Adversarial Network (CADN) \cite{long2018conditional}, Cycle-Consistent Adversarial Domain Adaptation (CyCADA) \cite{pmlr-v80-hoffman18a}, Re-weighted Adversarial Adaptation Network (RAAN) \cite{chen2018re}, Local Feature Patterns for Domain Adaptation (LFPDA) \cite{wen2019exploiting}. We follow standard evaluation protocols of unsupervised domain adaptation as in \cite{long2017deep}. For our model, we report performances with uncertainty estimated with entropy and variance formulations, denoted as \emph{Our(Entro)} and \emph{Our(Var)}, respectively.

\subsection{Implementation Details}

\paragraph{CNN Architectures.}

 For digit classification datasets, we use the same architecture as in ADDA \cite{tzeng2017adversarial}. All digit images are resized to $28\times28$ for fair comparisons.

On the \emph{Office-31} and the \emph{Office-home} datasets, we finetune the AlexNet pre-trained from the ImageNet. Following the DANN \cite{ganin2016domain}, a bottleneck layer \emph{fcb} with 256 units is added after the \emph{fc7} layer for adaptation. We adopt the same image random flipping and
cropping strategy as in JAN \cite{long2017deep}.

\begin{table}[t]
\small
\begin{center}
\resizebox{0.99\columnwidth}{!}{
\begin{tabular}{cccccc}
\toprule
 Method & SVHN$\rightarrow$MNIST & MNIST$\rightarrow$USPS & USPS$\rightarrow$MNIST  &Avg\\
\midrule
ADDA        &$76.0 \pm 1.8 $  &$89.4 \pm 0.2  $&$90.1 \pm 0.8 $   &$85.2$\\
RAAN      &$89.2 $ &$89.0 $&$92.1 $ &$90.1$ \\
LFPDA        &$86.9 \pm 0.5 $  &$92.2 \pm 0.4  $&$92.5 \pm 0.3 $   &$90.5$\\
CyCADA     &$90.4 \pm 0.4 $  &$95.6 \pm 0.2  $&$96.5 \pm 0.1 $   &$94.2$ \\
CDAN-M     &$89.2         $  &$\textbf{96.5} $&$97.1         $   &$94.3$\\
Ours(Var)      &$80.3   \pm 0.7      $  &$93.5 \pm 0.4 $&$94.7     \pm 0.3    $   &$89.5$\\
Ours(Entro)&$\textbf{91.5} \pm 0.3$&$95.7\pm 0.4$&$\textbf{98.1}\pm 0.2$&$\textbf{95.1}$\\
\bottomrule
\end{tabular}}
\normalsize
\caption{Accuracy ($\%$) of unsupervised domain adaptation on digits recognition tasks.}
\end{center}
\end{table}

\begin{table}[t]
\begin{center}
\small
\resizebox{0.99\columnwidth}{!}{
\begin{tabular}{cccccc}
\toprule
 Method & A$\rightarrow$W  &  A$\rightarrow$D  &  W$\rightarrow$A  &  D$\rightarrow$A    &Avg\\
\midrule
AlexNet     &$61.6 \pm 0.4 $  &$63.8 \pm 0.5  $&$49.8 \pm 0.4 $   &$51.1 \pm 0.6$ & $56.6$  \\
DANN        &$73.0 \pm 0.5 $  &$72.3 \pm 0.3  $&$51.2 \pm 0.5 $   &$53.4 \pm 0.4$  & $62.5$  \\
ADDA         &$73.5 \pm 0.6 $  &$71.6 \pm 0.4  $&$53.5 \pm 0.6 $   &$54.6 \pm 0.5$  & $63.3$  \\
LFPDA         &$75.2 \pm 0.3 $  &$72.1 \pm 0.5  $&$54.2 \pm 0.5 $   &$56.9 \pm 0.5$  & $64.6$  \\
JAN        &$74.9 \pm 0.3 $  &$71.8 \pm 0.2  $&$55.0 \pm 0.4 $   &$\textbf{58.3} \pm 0.3$  & $65.0$  \\
CDAN-M          &$78.3 \pm 0.2 $  &$76.3 \pm 0.1  $&$\textbf{57.3} \pm 0.3 $   &$57.3 \pm 0.2$  & $67.3$  \\
Ours(Entro)&$\textbf{78.9}\pm 0.4$&$\textbf{77.8} \pm 0.3$&$56.6 \pm 0.5$&$57.4 \pm 0.4$&$\textbf{67.7}$\\
\bottomrule
\end{tabular}}
\normalsize
\caption{Accuracy ($\%$) on the \emph{Office31} dataset for unsupervised domain adaptation.}
\end{center}
\end{table}

\paragraph{Hyper-parameters.}
To enable stable training, we progressively increase the importance of the adaptation loss and set $\lambda_{adv}= \frac{2}{1+\exp(\gamma \cdot m)}-1$, where $\gamma=-10$ and $m$ denotes the training progress ranging from 0 to 1. We use a similar hyper-parameter selection strategy as in DANN, called reverse validation. We set $\lambda_u=0.25 \lambda_{adv}$ to ensure uncertainty reduction. With $\tau=1.5$, we forward each sample $T=12$ times to obtain prediction uncertainty. We set $t_u=0.2$, for adaptation loss re-weighting, and $\tau_c=1.8$ for source classification loss. We dropout all fully-connected layers with a dropout ration $q=0.5$. Improvements are not observed with further dropout on convolution layers.

\subsection{Results}
The results on the digit recognition task are shown in Table 1. \emph{Our(Entro)} achieves the best performances on most of the tasks. The CyCADA align features at both pixel-level and feature-level. RAAN alleviates conditional distribution shift by matching label distributions. CADN-M attempts to learn domain-invariant interactions between learned features and classifier through conditional adversarial learning. On these tasks, the plenty of source labels prevents the low-capacity \emph{LeNet}-like model from overfitting the source labels, thus the advantages of our method over DAAN and CADN-M mainly come from uncertainty discrepancy minimization that alleviates the classifier bias.

\setlength{\tabcolsep}{1mm}
\begin{table*}[t]
\small
\begin{center}
\resizebox{\textwidth}{!}{
\begin{tabular}{cccccccccccccc}
\toprule[1pt]
    Method &Ar$\rightarrow$Cl&Ar$\rightarrow$Pr&Ar$\rightarrow$Rw&Cl$\rightarrow$Ar&Cl$\rightarrow$Pr&Cl$\rightarrow$Rw&Pr$\rightarrow$Ar&Pr$\rightarrow$Cl&Pr$\rightarrow$Rw&Rw$\rightarrow$Ar &Rw$\rightarrow$Cl&Rw$\rightarrow$Pr&Avg\\
    \midrule
AlexNet  &$26.3$&$32.6$&$41.3$&$22.1$&$41.7$&$42.1$&$20.5$&$20.3$&$51.1$&$31.0$&$27.9$&$54.9$&$34.3$\\
DANN              &$36.4$&$45.2$&$54.7$&$35.2$&$51.8$&$55.1$&$31.6$&$39.7$&$59.3$&$45.7$&$46.4$&$65.9$&$47.3$\\
JAN              &$35.5$&$46.1$&$57.7$&$36.4$&$53.3$&$54.5$&$33.4$&$40.3$&$60.1$&$45.9$&$47.4$&$67.9$&$48.2$\\
CDAN-M        &$38.1$&$50.3$&$60.3$&$\textbf{39.7}$&$56.4$&$57.8$&$\textbf{35.5}$&$43.1$&$\textbf{63.2}$&$48.4$&$48.5$&$71.1$&$51.0$\\
Ours(Entro)        &$\textbf{40.3}$&$\textbf{51.6}$&$\textbf{61.5}$&$37.9$&$\textbf{58.0}$&$\textbf{58.6}$&$33.6$&$\textbf{45.9}$&$61.8$&$\textbf{50.1}$&$\textbf{50.9}$&$\textbf{71.7}$&$\textbf{51.8}$\\
    \bottomrule[1pt]
\end{tabular}}
\normalsize
\caption{Accuracy ($\%$) on the \emph{Office-home} dataset for unsupervised domain adaptation.}
\end{center}
\end{table*}

\setlength{\tabcolsep}{3.5pt}

\begin{figure*}[ht]
\begin{center}
\subcaptionbox{Non-adapted}{\includegraphics[width=0.14\textwidth]{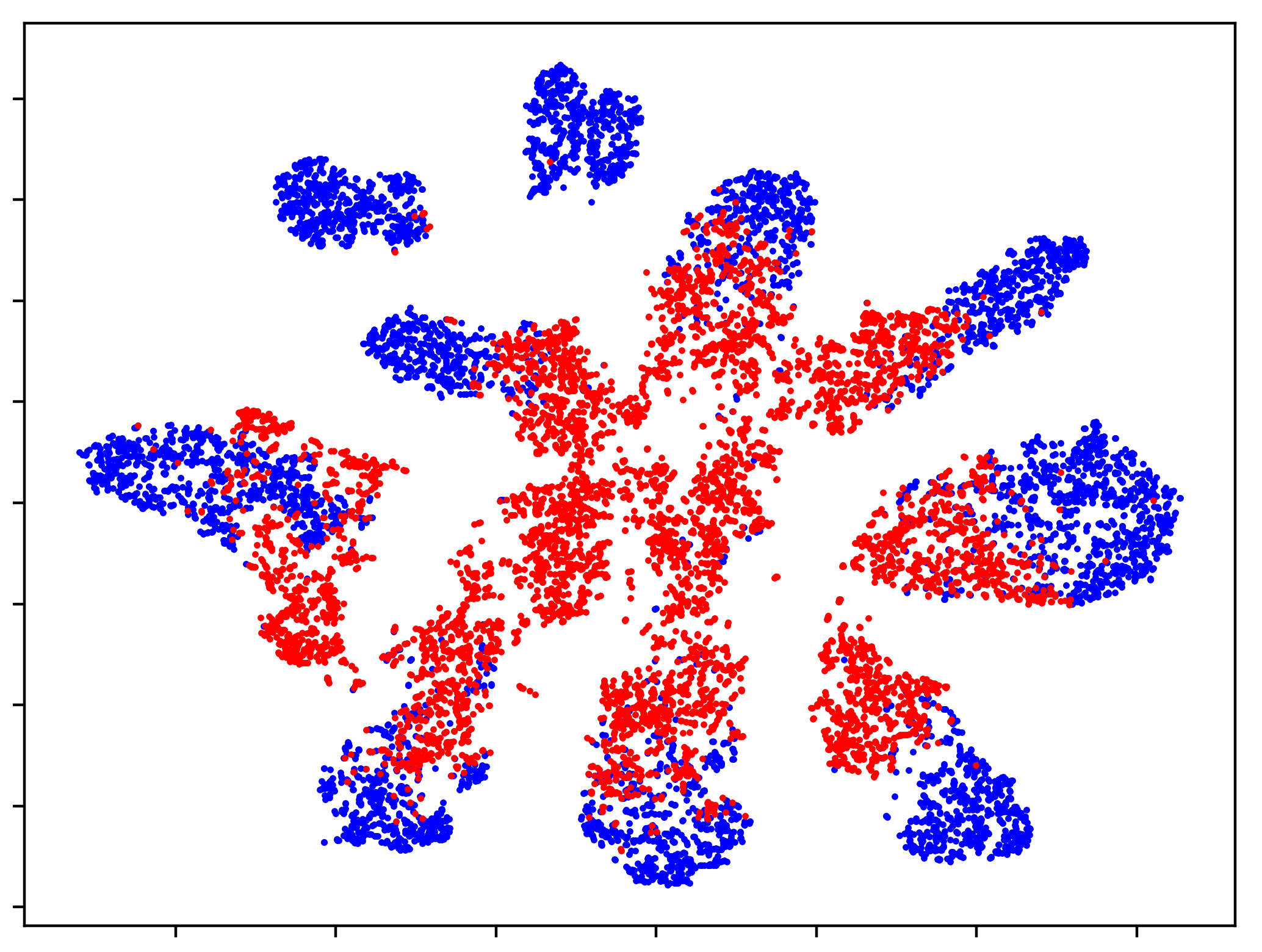}}\label{sample-table}%
\hfill
\subcaptionbox{DANN}{\includegraphics[width=0.14\textwidth]{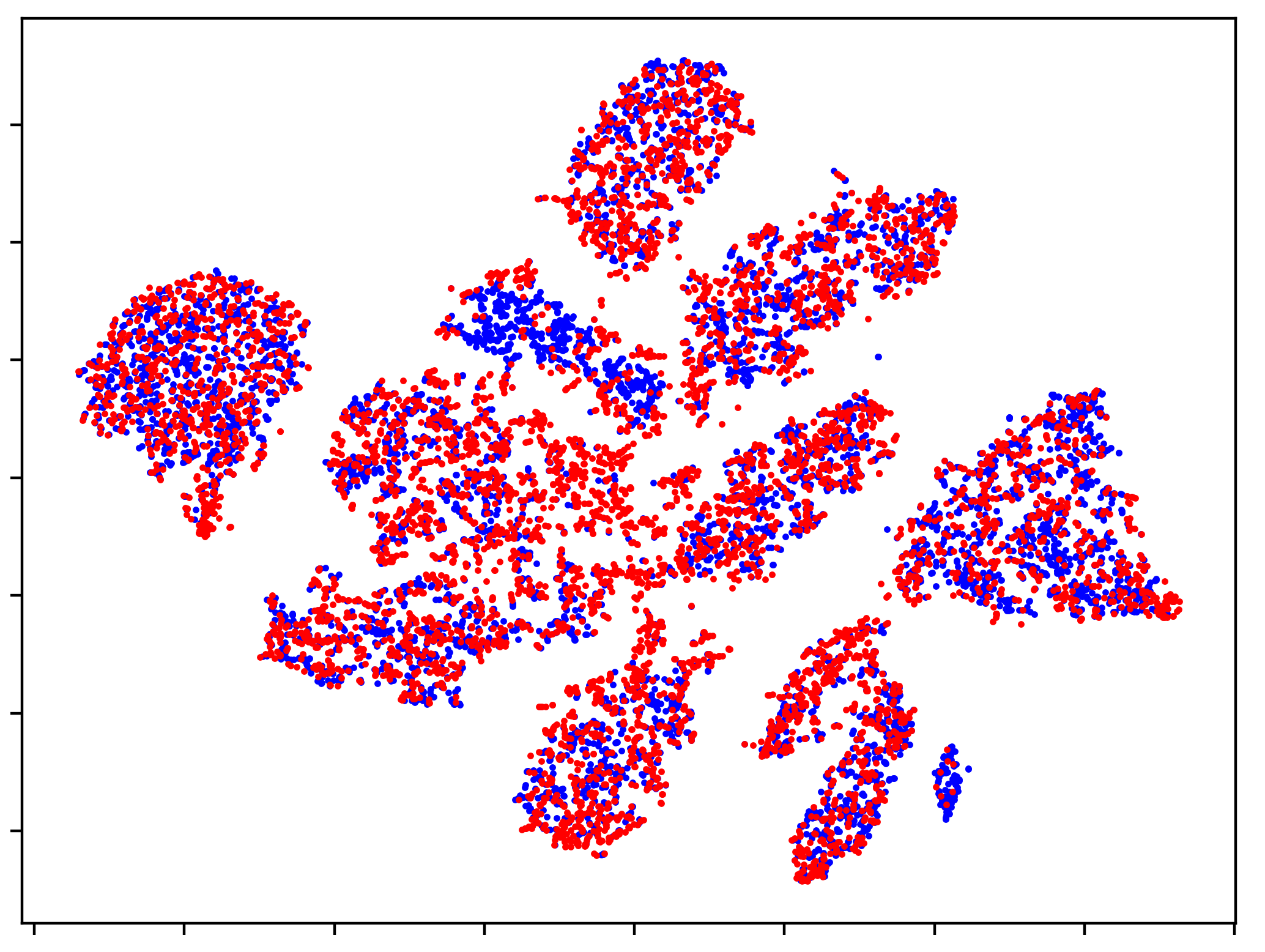}}
\hfill
\subcaptionbox{Ours(Entro)}{\includegraphics[width=0.14\textwidth]{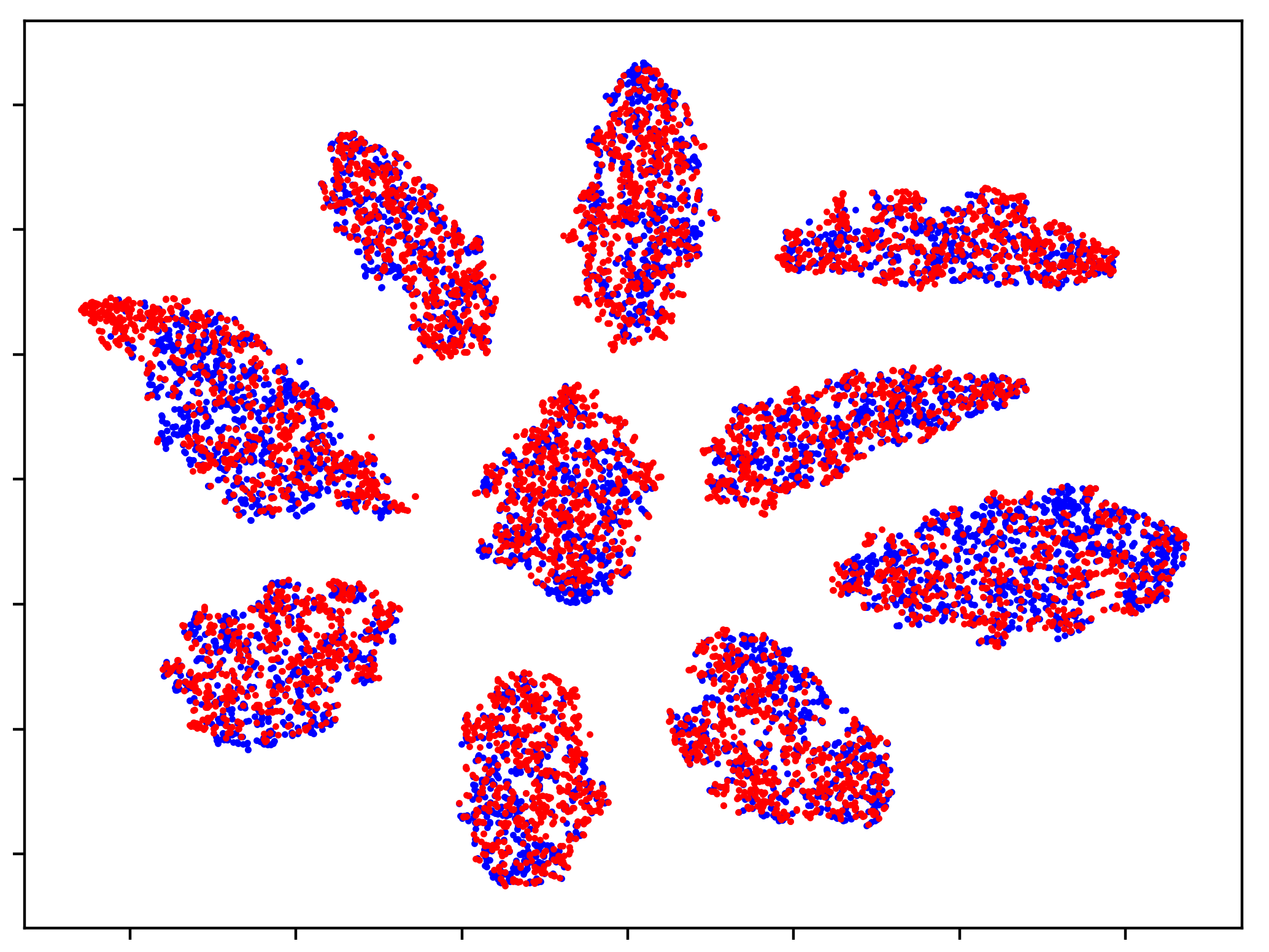}}%
\hfill
\subcaptionbox{DANN}{\includegraphics[width=0.14\textwidth]{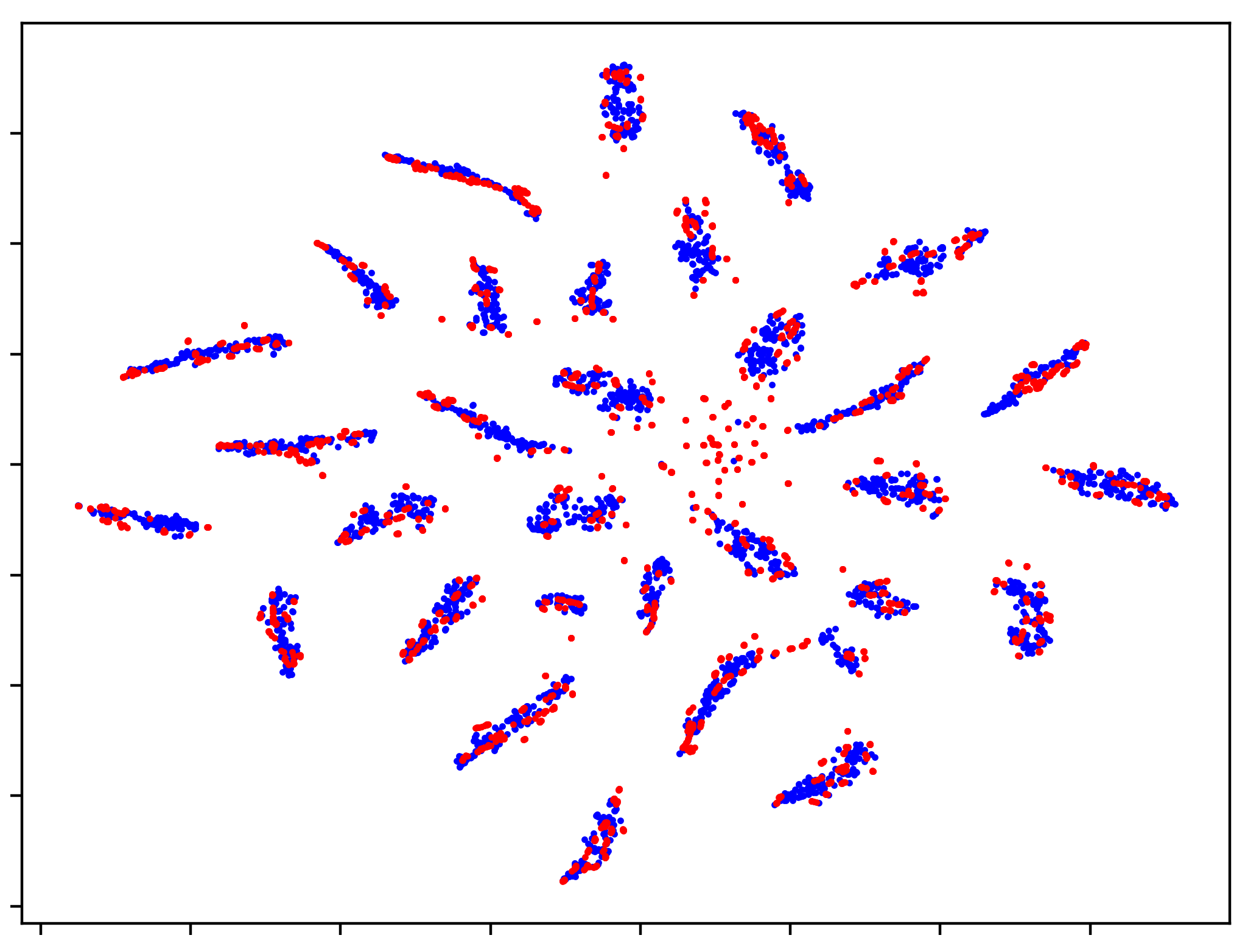}}%
\hfill
\subcaptionbox{Ours(Entro)}{\includegraphics[width=0.14\textwidth]{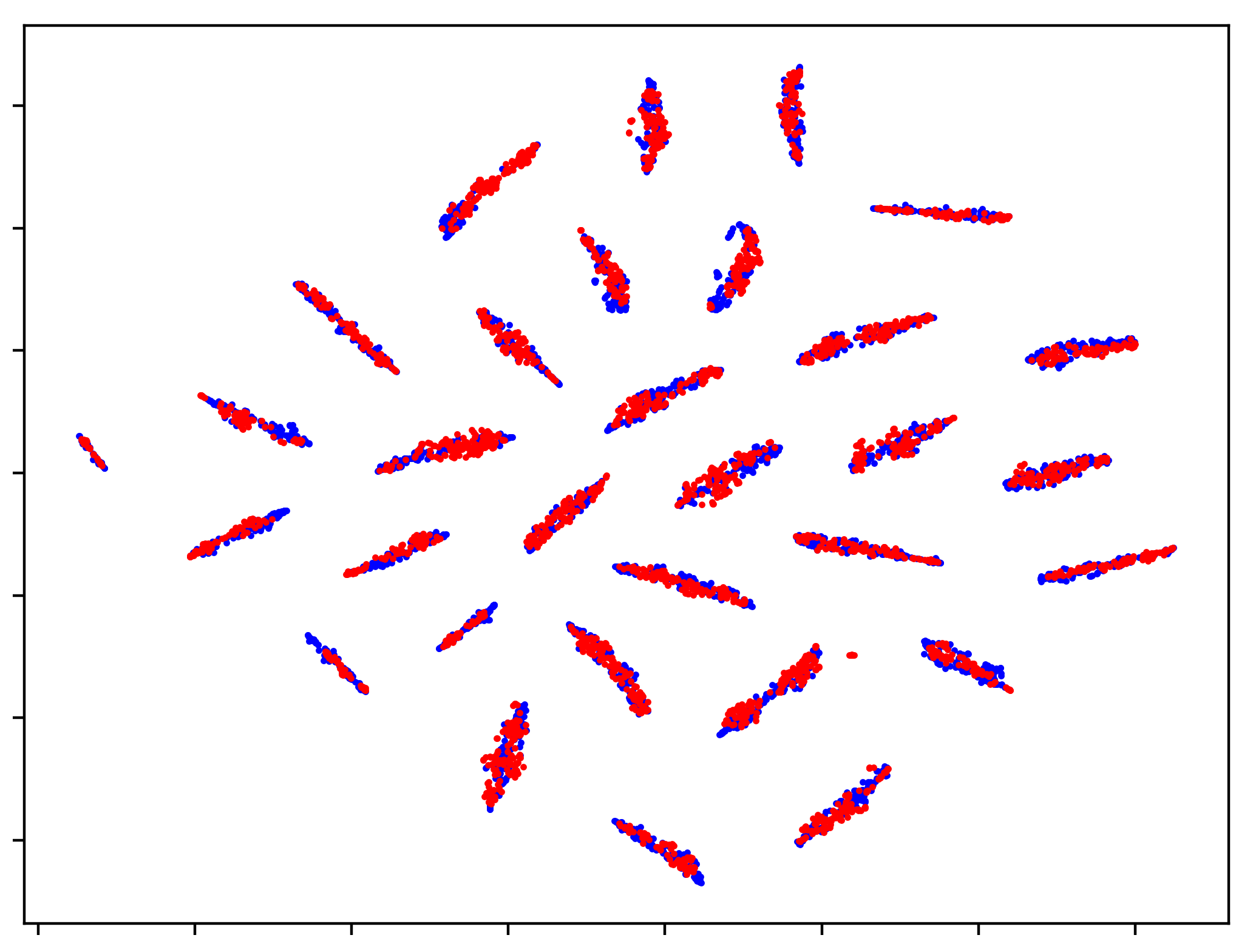}}%
\caption{The t-SNE visualizations of features on the USPS$\rightarrow$MNIST and A$\rightarrow$D tasks (blue: source; red: target). (a) is trained without adaptation; (b) and (d) is trained with the typical adversarial domain adaptation method DANN. (c) and (e) adapted using our method. Our method significantly reduces the marginal discrepancy while with much less boundary target features comparing to DANN (best viewed in color).}
\end{center}
\end{figure*}

\emph{Our(Entro)} consistently outperforms \emph{Our(Var)}. The distinct performance gap can be explained as follows. The entropy captures the cross-category probability spread of the prediction while the variance measures the deviation of prediction probabilities around the mean. The entropy uncertainty is more sensitive to the multi-peak probability spread across different categories. During training, the output probabilities of unmatched or boundary target samples usually cluster around two or more peaks, namely uncertain among several neighboring categories. In this case, the variance measure would obfuscate this multi-peak information. In the following, we only report the performances of \emph{Our(Entro)}.

\begin{table}[t]
\begin{center}
\resizebox{0.99\columnwidth}{!}{
\begin{tabular}{cccccc}
    \toprule[1pt]
    Method &A$\rightarrow$W & A$\rightarrow$D  & W$\rightarrow$A  &  D$\rightarrow$A  &  Avg\\
    \midrule
    AlexNet     &$58.2 (60.4) $&$60.4 (61.5)  $&$47.3 (45.8)  $&$49.8 (49.3)  $&$53.9 (54.3) $\\
    DANN        &$65.1 (70.7) $&$60.6 (72.5)  $&$42.9 (46.9) $&$42.1 (40.3)  $&$52.7 (57.6) $\\
    MADA         &$70.8(-) $&$69.6 (-)  $&$54.4 (-)  $&$54.2 (-)  $&$62.3 (-) $\\
    Ours(Entro) &$\textbf{73.4} (\textbf{76.2})  $&$\textbf{74.6} (\textbf{76.5}) $&$\textbf{55.5} (\textbf{54.8})$  &$\textbf{55.9} (\textbf{47.5})$&  $\textbf{64.9} (\textbf{63.8})$\\
    \bottomrule[1pt]
    \end{tabular}}%
\normalsize
\caption{Accuracy ($\%$) on the \emph{Office31} dataset with \emph{31$\rightarrow$25} and \emph{ 25$\rightarrow$25(+6)} adaptation tasks. For the \emph{25$\rightarrow$25(+6)} task, the extra 6 classes are treated as noisy images. In the table, $a$ in $a(b)$ denotes results of \emph{31$\rightarrow$25}, and $b$ denotes results of \emph{ 25$\rightarrow$25(+6)}. }
\end{center}
\end{table}

\begin{figure}[ht]
\vspace{-0.3cm}
\begin{center}
\subcaptionbox{USPS$\rightarrow$MNIST}{\includegraphics[width=0.24\textwidth]{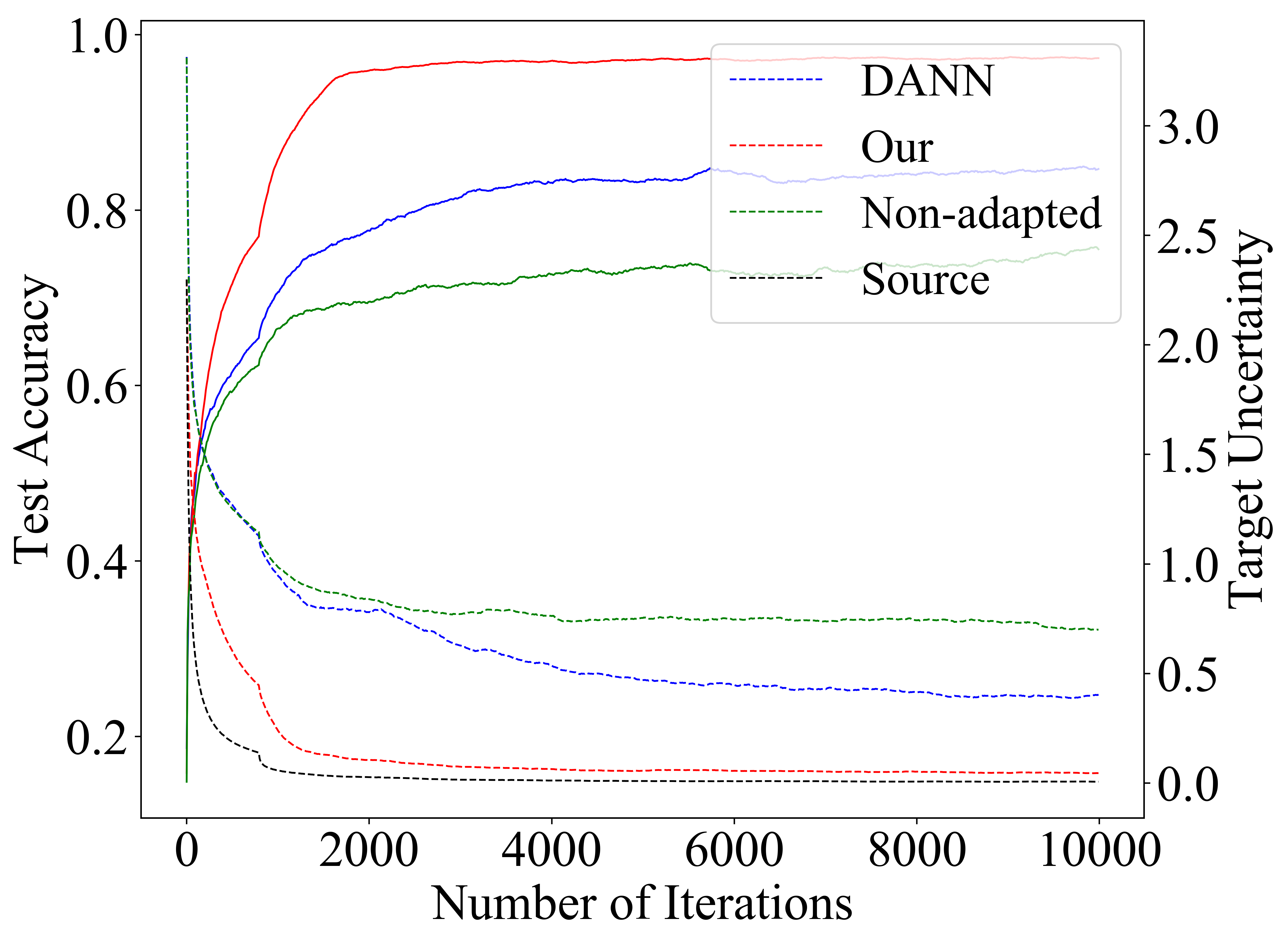}}%
\hfill
\subcaptionbox{A$\rightarrow$D}{\includegraphics[width=0.24\textwidth]{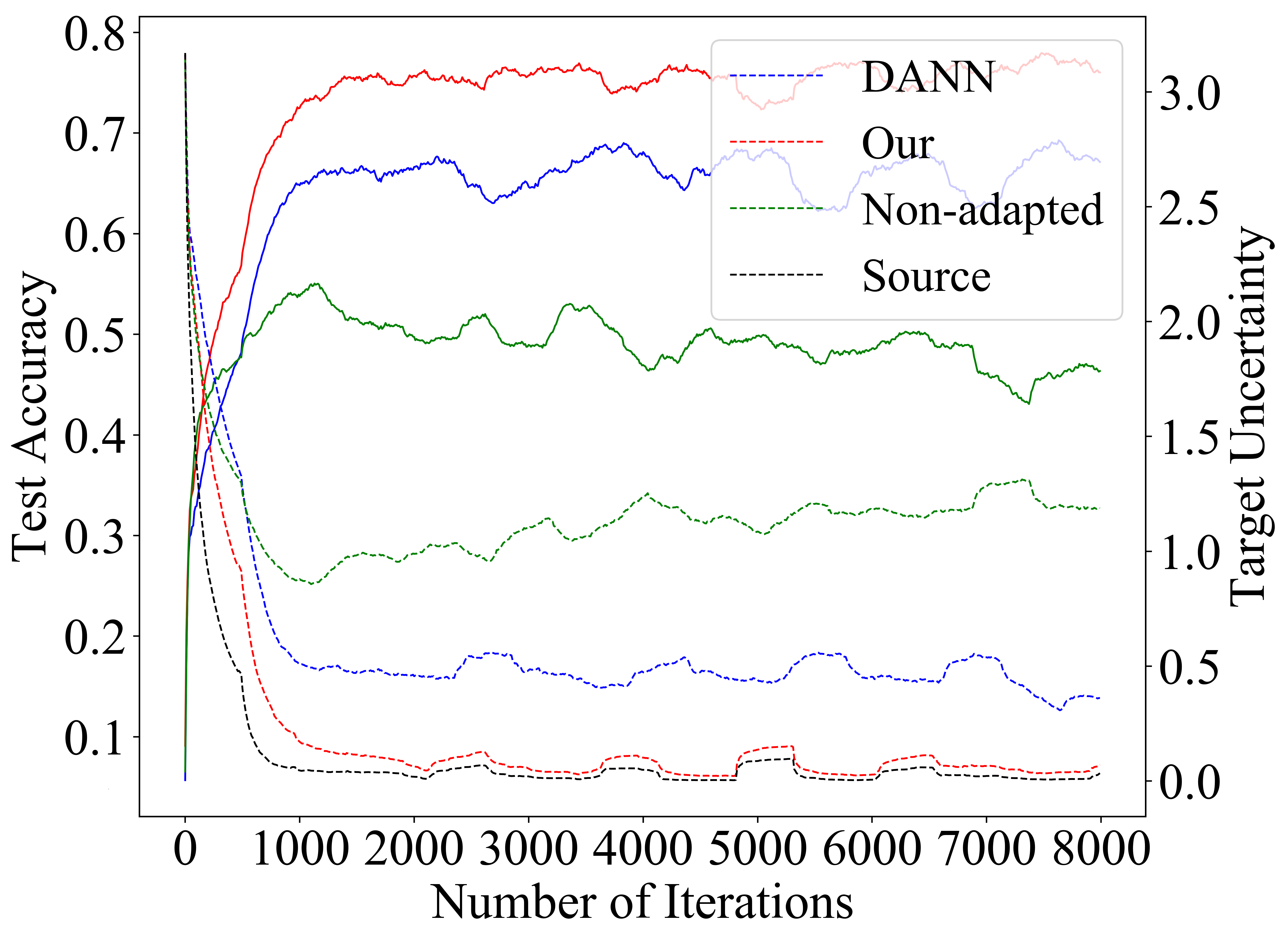}}%
\caption{Comparisons of target test accuracy and uncertainty on the USPS$\rightarrow$MNIST task and A$\rightarrow$D task of the \emph{Office31} dataset (dashed line: uncertainty; solid line: target test accuracy).  }
\end{center}
\vspace{-0.3cm}
\end{figure}

Performances on the \emph{Office-31} and \emph{Office-home} datasets are reported in Table 2 and Table 3, respectively. Again, our model achieves the best performances on most of the tasks. Due to the smaller size of the labeled source dataset and the huge capacity of the AlexNet, the models easily overfit the source labels while being jointly trained to reduce the marginal distribution discrepancy. The overfitting harms the transferability of the aligned features, resulting in learning trivial features for the target domain. In this case, our model alleviates this problem by jointly enforcing feature alignment and classifier prediction consistency.

\paragraph{Negative Transfer.}
Negative transfer happens when features are falsely aligned and domain adaptation causes deteriorated performances. Existing marginal distribution matching methods easily induce negative transfer when the marginal distributions between source and target are inherently different, {\it e.g.}, the source domain is smaller or larger than the target. We conduct experiments on the \emph{Office-31} dataset with the \emph{31$\rightarrow$25} task by removing 6 classes from the target, and the \emph{25$\rightarrow$25(+6)} task by treating 6 extra target classes as noise images. We compare our method with DANN and \emph{MADA} \cite{pei2018multi} which is showed effective on alleviating negative transfer. The results are reported in Table 4. It is seen that DANN suffers obvious negative transfer on the \emph{31$\rightarrow$25} task. The effectiveness of our method on alleviating negative transfer is significant. Adaptive joint feature-uncertainty distribution matching encourages the model to mix source and target samples that best match with each other, thus alleviating the harmful effects of noisy samples.

\paragraph{Alignment Visualization.}
We visualize the source and target learned representations on the USPS$\rightarrow$MNIST and $A\rightarrow$D adaptation tasks using the t-SNE embedding \cite{maaten2008visualizing}. In Figure 3, we visualize features of non-adapted models, DANN and our adapted model. Compared with the non-adapted model, DANN significantly reduces marginal distribution shift. Our method effectively prevents generating unmatched target samples that lie close to the decision boundary of the classifier and tend to be incorrectly classified.

\paragraph{Convergence and Uncertainty.}
In Figure 4, we show the convergence (test accuracy) and target uncertainty of the non-adapted model, DANN, and our model, on the USPS$\rightarrow$MNIST and A$\rightarrow$D tasks. As we can see, DANN adaptation effectively reduces target prediction uncertainty (source uncertainty is assured to be low) and improves target test accuracy. Our model further significantly reduces the discrepancy between source and target prediction uncertainty. The nearly synchronous increase of target accuracy and decrease of cross-domain prediction uncertainty discrepancy further indicates that uncertainty matching alleviates domain-shift and improves domain adaptation.

\section{ Conclusions}
We have proposed a novel and effective approach for joint-distribution matching by exploiting prediction uncertainty. To achieve this, we adopt a Bayesian neural network to model prediction uncertainty. Unlike most of existing deep domain-adaptation methods that only reduce marginal feature-distribution shift, the proposed method additionally alleviates conditional distribution shift lingering in the last classifier. Experimental results verify the advantages of the proposed method over state-of-the-art unsupervised domain-adaptation approaches. More interestingly, we also have shown that the proposed method can effectively alleviate negative transfer in domain adaptation.

\section*{Acknowledgments}

This work is supported by the Zhejiang Provincial Natural Science Foundation (LR19F020005), National Natural Science Foundation of China (61572433, 31471063, 31671074) and thanks for a gift grant from Baidu inc. Also partially supported by the Fundamental Research Funds for the Central Universities.

\bibliographystyle{named}
\bibliography{Uncer_UDAbib}

\end{document}